\documentclass[a4paper, 10pt, conference]{IEEEtran}
\IEEEoverridecommandlockouts
\usepackage{cite}
\usepackage{amsmath,amssymb,amsfonts}
\usepackage{graphicx}
\usepackage{textcomp}
\usepackage{caption}
\usepackage{subcaption}

\usepackage{algorithm}
\usepackage{booktabs}
\usepackage{caption}
\usepackage{subcaption}
\usepackage{multirow}
\usepackage{algorithm}

\usepackage[english]{babel}
\usepackage[utf8]{inputenc}
\usepackage[noend]{algpseudocode}
\DeclareMathOperator*{\argmax}{argmax} 

\usepackage{xcolor}
\def\BibTeX{{\rm B\kern-.05em{\sc i\kern-.025em b}\kern-.08em
    T\kern-.1667em\lower.7ex\hbox{E}\kern-.125emX}}
\begin{document}

\title{Feature Decoupling in Self-supervised Representation Learning for Open Set Recognition \\}

\author{\IEEEauthorblockN{ Jingyun Jia}
\IEEEauthorblockA{\textit{Department of Computer Engineering and Sciences} \\
\textit{Florida Institute of Technology}\\
Melbourne, US \\
Email: jiaj2018@my.fit.edu}
\and
\IEEEauthorblockN{Philip K. Chan}
\IEEEauthorblockA{\textit{Department of Computer Engineering and Sciences} \\
\textit{Florida Institute of Technology}\\
Melbourne, US \\
Email: pkc@cs.fit.edu}
}

\maketitle

\begin{abstract}
Assuming unknown classes could be present during classification, the open set recognition (OSR) task aims to classify an instance into a known class or reject it as unknown. In this paper, we use a two-stage training strategy for the OSR problems. In the first stage, we introduce a self-supervised feature decoupling method that finds the content features of the input samples from the known classes. Specifically, our feature decoupling approach learns a representation that can be split into content features and transformation features. In the second stage, we fine-tune the content features with the class labels. The fine-tuned content features are then used for the OSR problems. Moreover, we consider an unsupervised OSR scenario, where we cluster the content features learned from the first stage. To measure representation quality, we introduce intra-inter ratio (IIR). Our experimental results indicate that our proposed self-supervised approach outperforms others in image and malware OSR problems. Also, our analyses indicate that IIR is correlated with OSR performance.

\end{abstract}

\begin{IEEEkeywords}
Open set recognition; self-supervised learning; unsupervised learning; representation learning
\end{IEEEkeywords}

\section{introduction}

As classification techniques have achieved great success in various fields in research and industry, most traditional classification problems focus on the known classes. However, collecting samples exhausting all classes in the real world is difficult. This problem is referred as Open Set Recognition (OSR) \cite{bendale2016towards}. OSR attempts to handle the known classes that already exist in the training set and the unknown classes that are absent from the training set. Hence, for a multinomial classification problem, an OSR task normally involves two objectives: to classify the known classes and reject the unknown class. These two objectives help us building a more robust system than a traditional classifier, such system defines a more realistic scenario and benefits the applications like face recognition \cite{DBLP:journals/cviu/OrtizB14}, malware classification \cite{DBLP:journals/corr/abs-1802-04365} and medical diagnoses \cite{DBLP:conf/ipmi/SchleglSWSL17}.

In this paper, we introduce a two-stage learning strategy for the OSR problem. The first stage extracts the content features via a self-supervised learning approach. The majority of the self-supervised learning methods focus on designing various pre-text tasks \cite{DBLP:conf/iclr/GidarisSK18, DBLP:conf/icml/ZbontarJMLD21,DBLP:conf/icann/JiaC22}. These pretext tasks usually aim to learn content features, which implicitly try to remove the transformation information. We hypothesize that explicitly learning separate content and transformation features can improve the content features. During the first (pre-training) stage, we introduce a feature decoupling approach to extract the content features irrelevant to transformation information. The proposed approach decouples the representations through content reconstruction and transformation prediction tasks. A learned representation in our network is a concatenation of content features and transformation features. The content features are irrelevant to the transformations. Thus, the content features of the different views from the same original input should be the same. We achieve this goal by reconstructing the different views to their original form. Furthermore, the transformation features should contain the discriminative information on the transformation types. Thus, we introduce the transformation labels inside the input transformation module and build an auxiliary transformation classifier on top of the transformation features. After explicitly learning separate content and transformation features, in the second stage, we fine-tune the content features with the provided class labels and discard the transformation features. We look into two different supervised loss functions: classification loss and representation loss. The classification loss, such as cross-entropy loss, is applied to the decision layer to lower the classification error. The representations loss, such as triplet loss \cite{DBLP:conf/cvpr/SchroffKP15}, is applied directly to the representation layer to decrease the intra-class spread and increase inter-class separation. In the case of classification loss, we connect a content classifier to the content features and fine-tune the network with a content classification loss. In the case of representation loss, we apply the loss function directly to the content features. Finally, the fine-tuned content features are used for the OSR tasks. We also consider an unsupervised learning scenario in the second stage, where the known class labels are unavailable. In this case, we cluster the content features learned in the first stage to find the potential classes.

Our contribution includes: first, we design a two-stage training strategy for the OSR tasks. Among these, we propose a feature decoupling approach to extract the content features that are irrelevant to the transformation information. Second, we extend our approach to the unsupervised scenario in OSR. Third, to evaluate the quality of learned representations of the pre-training and fine-tuning stages, we propose intra-inter ratio (IIR) and show that it is correlated to OSR performance. Lastly, we experiment with different loss functions with image and malware datasets. The results indicate that our proposed self-supervised learning method is more effective than other approaches in OSR.

We organize this paper as follows. Section 2 reviews the related research works on OSR systems and self-supervised learning approaches. Section 3 presents a two-stage training strategy for learning content features and introduces how to use the learned content features to perform the OSR tasks. In Section 4, we evaluate our proposed approach through experiments on different types of datasets and also compare the experimental results with other approaches.
\section{related work}

\subsection{Open Set Recognition (OSR)}
An OSR task has two objectives: classify the known classes and recognize the unknown class while the unknown class is absent from the training set. Recent research works have developed various approaches to deal with the absence of the unknown class, based on which, we divide these works into three categories.

The approaches in the first category borrow other data samples used as the unknown class. Dhamija et al.\ \cite{DBLP:conf/nips/DhamijaGB18} utilize the differences in feature magnitudes between known classes. They borrow unknown samples as part of the objective function. Shu et al.\ \cite{shu2018odn} propose ODIN, which uses Emphasis Initialization and Allometry Training to initialize and incrementally train the new predictor. The approach borrows a few novel samples to fine-tune the model. 
The approaches in the second category generate additional data as the unknown class. Ge et al.\ \cite{DBLP:conf/bmvc/GeDG17} introduce a conditional GAN to generate some unknown samples followed by an OpenMax classifier. Yu et al.\ \cite{DBLP:conf/ijcai/YuQLG17} propose an ASG framework. It uses the min-max strategy from GANs to generate data around the decision boundary between known and unknown samples as unknown. 
Lee et al.\ \cite{DBLP:conf/iclr/LeeLLS18} generate ``boundary'' samples in the low-density area of in-distribution acting as unknown samples. Zhou et al.\ \cite{DBLP:conf/cvpr/ZhouYZ21} present PROSER to generate the unknown samples from a mixup of the representations of two different known samples and use them in training.
The third category of OSR approaches does not borrow nor generate additional data as the unknown class in training. Bendale and Boult \cite{DBLP:conf/cvpr/BendaleB16} propose OpenMax for the OSR problems. OpenMax adapts Meta-Recognition concepts to the activation patterns in the representation layer of the network and then estimates the probability of an input being from an unknown class. 
Hassen and Chan\cite{DBLP:journals/corr/abs-1802-04365} propose ii loss for open set recognition. It first finds the representations for the known classes during training and then recognizes an instance as unknown if it does not belong to any known classes. Jia and Chan \cite{DBLP:conf/icann/JiaC21} propose MMF as a loss extension to further separate the known and unknown representations for the OSR problem. CROSR in \cite{DBLP:conf/cvpr/YoshihashiSKYIN19} trains networks for joint classification and reconstruction of the known classes to combine the learned representation and decision in the OSR task. Moreover, Perera et al.\ \cite{DBLP:conf/cvpr/PereraMJMWOP20} adopt a self-supervision framework to force the network to learn more informative features when separating the unknown class. Specifically, they used the autoencoder output as auxiliary features for the OSR task. Our proposed approach does not borrow nor generate additional unknown samples. Thus it falls into the third category.

\subsection{Self-supervised Learning}
Self-supervised learning is a subcategory of unsupervised learning. Unlike the traditional supervised learning approaches, which use human annotations as the guidance of the training process for the primary tasks, the self-supervision approaches learn representations via pretext tasks that are different from the primary tasks. Autoencoding is a simple example. It learns the representation by reconstructing the original input samples. Besides autoencoder, most research works introduce these pretext tasks by transforming the original inputs into different views. Recent works have applied a self-supervised learning approach to various domains. For example, in the image representation learning, Gidaris et al.\ \cite{DBLP:conf/iclr/GidarisSK18} propose RotNet, where the pretext task is predicting the rotation degree (e,g,. 0, 90, 180 and 270 degrees). To achieve better generalization ability, Feng et al.\ \cite{DBLP:conf/cvpr/FengXT19} decouples the rotation discrimination from instance discrimination. Chen et al.\ \cite{DBLP:conf/icml/ChenK0H20} propose SimCLR. It first generates positive and negative pairs between different views and proposes contrastive loss to increase the similarity between positive pairs meanwhile decrease the similarity between negative pairs. Zbontar et al.\ \cite{DBLP:conf/icml/ZbontarJMLD21} propose Barlow Twins, which generates the cross-correlation matrix between the representations of two views of the same input. Then, it tries to make this matrix close to the identity to reduce the redundancy in learned features. Moreover, Jia and Chan \cite{DBLP:conf/icann/JiaC22} propose DTAE, which reconstructs the different views back to their original forms. Thus, the representations of different views of the same input should be similar. 

Besides image representation learning, more recent works have extended self-supervision approaches to graph representation learning. For example, You et al. \cite{DBLP:conf/nips/YouCSCWS20} propose Graph contrastive learning (GraphCL), which extends the contrastive learning framework in SimCLR to the graph field. Specifically, it designs four types of transformations for the graph inputs: node dropping, edge perturbation, attribute masking, and subgraph sampling. Jia and Chan extend the DTAE framework to function call graphs (FCGs) by introducing two FCG transformations: FCG-shift and FCG-random. Moreover, Jin et al.\ \cite{DBLP:conf/ijcai/JinZL00P21} propose Pairwise Half-graph Discrimination (PHD). PHD first generates two augmented views based on local and global perspectives from the input graph. Then, the objective function maximizes the agreement between node representations across different views and networks. Our proposed feature decoupling approach extracts the content features from the learned features. Regardless of the transformation type, feature decoupling can be used for image and graph representation learning. 
\section{Approach}
\label{sec: approach}

In this section, we first describe a two-stage learning process to learn the representations of input samples. In the first stage (pre-training), we utilize a self-supervision approach to extract the low-level content features of the input samples. In the second stage (fine-tuning), we introduce two types of loss functions (classification loss and representation loss) to fine-tune the discriminative content features. Then, we present a recognition strategy for the OSR problem with the centroids of the fine-tuned content features. Moreover, we consider an unsupervised scenario for the second stage of learning, where the labels of known classes are unavailable. In this case, we cluster the learned content features with K-Means instead of fine-tuning them with class labels in the second stage. Moreover, cluster centroids are used in the recognition strategy for the OSR problem.

\subsection{Pre-training stage: self-supervised feature decoupling}
\label{sec: pre}
Self-supervised learning uses pretext tasks in the objectives, which generally incorporate the transformations of original input samples. Thus, the learned features contain two types of information for a transformed input sample: transformation unrelated content information and transformation information. The information solely related to transformation is introduced by pretext tasks, which are not practical for the downstream tasks. Therefore, we develop a feature decoupling method to separate these two types of information into transformation unrelated content features and transformation features. 

\begin{figure}
    \includegraphics[width=\linewidth]{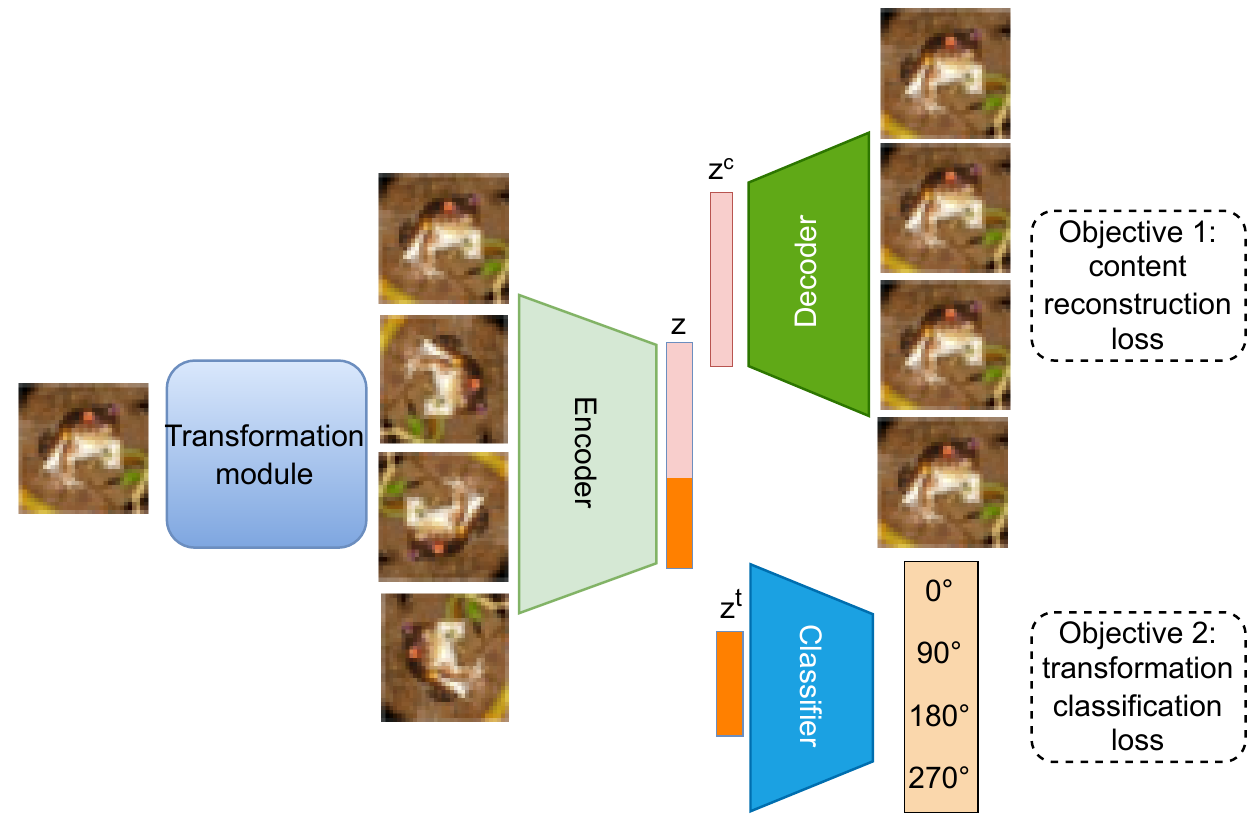}
  
                \caption{Illustration of proposed feature decoupling method. The transformation module transforms the original input samples into several correlation views. The encoder outputs decoupled content and transformation features. The content part is learned by reconstructing the transformed input samples back to their original forms, and the transformation part is learned by transformation classification. }
                \label{fig: pre-train}
\end{figure}

As shown in Figure \ref{fig: pre-train}, given an original input sample $x$, we use a transformation module $T$ augments the input $x$ with different several correlated views. The transformation module contains $M$ different transformations as $T = \{t_1, t_2, ...t_m\}$. In our example in Figure \ref{fig: pre-train}, the transformations are the rotations of input image samples by multiples of 90 degrees (0°, 90°, 180° and 270°) such that $M=4$. We denote the original input $x$ with transformation $t_j$ by $x_{j}$. i.e, $x_j = t_j(x)$. Then, a network-based encoder $f(\cdot)$ extracts the representation vector $z_j$ from transformed data example $x_j$, such that $z_j = f(x_j)$.  We suppose that the high level representation vector $z_{j}$ can be represented as $z_{j} = [z^c_{j}, z^t_{j}]$, where $z^c_{j}$ is content features of the transformed data example while $z^t_{j}$ is responsible for the transformation features. We apply two different objectives to decouple these two types of information. Specifically, we use a reconstruction decoder $g^c(\cdot)$ to learn the content features and a transformation classifier $g^t(\cdot)$ to extract the transformation related features.

\subsubsection{Learning content features}
The content features should be invariant to the transformations. In other words, for the same input sample, the content features should be invariant for all its transformed views. SimCLR \cite{DBLP:conf/icml/ChenK0H20} and Barlow Twins \cite{DBLP:conf/icml/ZbontarJMLD21} achieve such agreement by maximizing similarity of representations obtained from different transformed views of a sample in the latent space. DTAE \cite{DBLP:conf/icann/JiaC22} encourages the similarity of representations of different views by reconstructing them back to their original forms. Here, we apply a content reconstruction decoder on reconstructed transformed samples. As shown in the ``Objective 1'' in Figure \ref{fig: pre-train}, the input of the decoder is the content part of the representation, $z^c_{j}$. Moreover, instead of reconstructing the content features back to their transformed views, the decoder here ``reconstructs'' them to their original form before the transformation module.

Specifically, let $g^c(z^c_{j})$ denotes reconstruction from the content feature of the transformed view $x_{j}$, we use MSE (Mean Squared Error) loss to maximize the similarity of the reconstruction and the original input sample $x$:
\begin{equation}
    \mathcal{L}_\text{content} = \frac{1}{2}  \sum^{M}_{j=1}(x - g^c(z^c_{j}))^2
\label{eq: content}
\end{equation}
Where each of the data points has $M$ transformations, and there are $M$ times data points as the original input sample after the transformation module.

\subsubsection{Learning transformation features}

Towards the goal of extracting transformation features, we apply a classifier to predict the transformation classes introduced from the transformation module. As shown in the ``Objective 2'' in Figure \ref{fig: pre-train}, the input of the transformation classifier is the transformation part of the representation, $z^t_{j}$, and the output is the prediction logits of transformation classes, i.e., the rotation angles in our example. Formally, given the transformation part of the representation $z^t_{j}$, the classifier outputs the transformation prediction logit of the $i$-th class: $p_i(g^t(z^t_{j}))$. We use a softmax cross-entropy loss for the transformation classification and write the loss functions as: 

\begin{equation}
\mathcal{L}_\text{transformation} = -\log p_{i=j}(g^t(z^t_{j})),
\label{eq: trans}
\end{equation}
where $j$ is the ground truth transformation label of input sample $x_{j}$. In our example in Figure \ref{fig: pre-train}, the objective of the classifier is classifying four rotation types that introduced from the transformation module. 


\begin{algorithm}
\caption{Pre-training stage of feature decoupling in self-supervised representation learning}
 \hspace*{\algorithmicindent}\textbf{Input:} Training data and labels ($x$, $y$). \\
\hspace*{\algorithmicindent}\textbf{Output:} Encoder $f(\cdot)$,  content decoder $g^c(\cdot)$, and \\
\hspace*{\algorithmicindent} \hspace*{\algorithmicindent}\hspace*{\algorithmicindent}           transformation classifier $g^t(\cdot)$.
\begin{algorithmic}[1]
\State Random Initialize $f(\cdot)$,  $g^c(\cdot)$ and $g^t(\cdot)$;
\For{each transformation $j$}
\State $j, x_j \xleftarrow{} t_j(x)$
\EndFor
\For{epochs}
 \For{each transformation $j$}
  \State Extract the representation $z_j [z^c_j, z^t_j]$ from $f(x_j)$;
  \State Reconstruct the original input $x$ from $g^c(z^c_j)$; 
  \State Train $f(\cdot)$ and $g^c(\cdot)$ by Eq. \ref{eq: content}; 
 
  \State Classify the transformation label $j$ from $g^t(z^t_j)$; 
  \State Train $f(\cdot)$ and $g^t(\cdot)$ by Eq. \ref{eq: trans}; 
\EndFor
  \EndFor
\Return $f(\cdot)$, $g^c(\cdot)$, $g^t(\cdot)$.
\end{algorithmic}
 \label{algo: pre-train}
\end{algorithm}

\begin{figure*}
  \begin{subfigure}{\columnwidth} 
  \centering
  \includegraphics[width=0.8\linewidth]{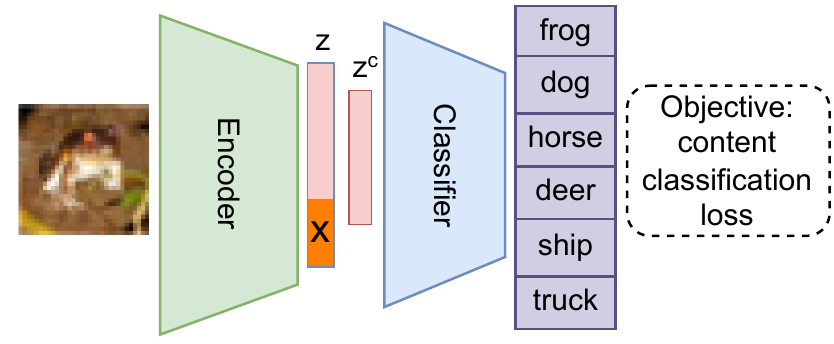}
  
                \caption{Fine-tuning with classification loss}
                \label{fig: fine-tune-cls}
        \end{subfigure}%
  \begin{subfigure}{\columnwidth} 
  \centering
  \includegraphics[width=0.8\linewidth]{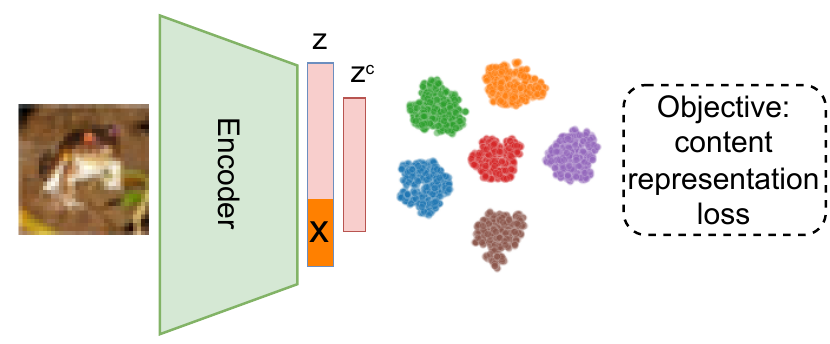}
                \caption{Fine-tuning with representation loss}
                \label{fig: fine-tune-rep}
        \end{subfigure}%
\caption{Illustration of two types of fine-tuning objectives. only the original input sample are used in the fine-tuning stage. The encoder and representation layer are inherited from the pre-training stage, then the content features ($z^c$, pink) are connected to (a) a classifier or (b) a representation loss function for fine-tuning. The transformation features (orange) are discarded.}
\label{fig: fine-tune}
\end{figure*}
In Algorithm \ref{algo: pre-train}, we summarize the overall pre-training stage of the proposed feature decoupling method. After initializing the networks in Line 1 and transforming the original training data, the network is trained using mini-batch stochastic gradient descent with backpropagation. During each epoch, Lines 7-8 train the encoder $f(\cdot)$ and content decoder $g^c(\cdot)$ under the guidance of content reconstruction loss function (Equation \ref{eq: content}) on the content part of the representation $z^c_j$. Next, Line 9 and Line 10 execute the training for the transformation part of the representation $z^t_j$ under the transformation classification loss function in Equation \ref{eq: trans}.

\subsection{Fine-tuning stage: supervised fine-tuning}
\label{sect: ft}

The self-supervised feature decoupling approach in the first stage attempts to find the low-level content features. We further fine-tune the content features learned in the first stage with the class labels. The fine-tuning stage incorporates the available class labels in the training data to discover the discriminative features between classes for class awareness. The loss functions of the fine-tuning network can be categorized into two types: classification and representation loss. 

The classification loss function requires a classifier connected to the representation layer, which is applied to the output logits in the decision layer. One of the widely used classification loss functions is cross-entropy loss. Figure \ref{fig: fine-tune-cls} illustrates the network architecture of using the classification loss in the fine-tuning step. Compared with the pre-training step, the fine-tuning step only uses the original training data. The training data is passed through the encoder learned in the pre-training step. Moreover, instead of connecting to a content decoder, the content part of the representation connects to a classifier that outputs class logits. The classification loss is applied to this output to lower the classification error. 

Unlike the classification loss, the representation loss functions do not require a classifier. They constrain the representation layers directly, such as triplet loss \cite{DBLP:conf/cvpr/SchroffKP15}. 
Figure \ref{fig: fine-tune-rep} illustrates how we incorporate the representation loss in the fine-tuning stage. After passing the input data through the pre-trained encoder, we extract the decoupled representations. Then, instead of a content decoder or a classifier, the content part of the representation is directly constrained by the representation loss function. After fine-tuning the encoder with the labeled dataset, we calculate centroid $u_k$ of class $k$ based on the content features:

\begin{equation}
    \mu_k = \frac{1}{N_k} \sum^{N_k}_{i=1} z^c_i,
\label{eq: centroid}
\end{equation}
where $N_k$ is the number of training instances in class $k$. During the inference time, we only use the content features $z^c$ to represent the input sample for the OSR tasks.

\subsection{Open set recognition}
\label{sec: osr}
After the second stage, we obtain the encoder and centroids of all the known classes. We have two problems to solve for an OSR task: classifying the known classes and identifying the unknown class. If we have $K$ known classes, given the content features $z^c$ of test sample $x$, we define the outlier score as the Euclidean distance to its closest centroid:
\begin{equation}\label{eq:outlier}
outlier\_score(x) = \min_{1\leq{k}\leq{K}}\|\mu_k-z^c\|_{2}^2
\end{equation}

In this work, the outlier threshold $t$ is the 99 percentile of the outlier score in ascending order. A test sample is recognized as unknown if an outlier score exceeds the selected threshold. Otherwise, we use a class probability $P(y=k|x)$ to decide the test sample belongs to which known class. For the fine-tuning network with classification loss, we use the output probability in the decision layer as $P(y=k|x)$. For the fine-tuning network with representation loss, we calculate $P(y=k|x)$ as:

\begin{align}\label{eq: prob}
P(y=k | x) = \frac{e^{-\|\mu_k-z\|_{2}^2}}{\sum_{k=1}^{K}e^{-\|\mu_k-z\|_{2}^2}}
\end{align}

And the test sample is classified as the known class with the highest class probability. 

\begin{equation}\label{eq:osr}
    \hat{y}=
    \begin{cases}
      unknown,& \text{if$\ outlier\_score(x) > t$} \\
      \argmax\limits_{1\leq{k}\leq{K}} P(y=k | x),& \text{otherwise}
    \end{cases}
\end{equation}

\subsection{ Extension to unsupervised OSR}
\label{sect: cluster}
When the class labels of the known classes are unavailable, the problem becomes an unsupervised OSR problem. The unsupervised OSR aims to identify whether an instance is from the known data distributions or an unknown data distribution without known class labels. In this scenario, after self-supervised pre-training (Sec. \ref{sec: pre}), instead of supervised fine-tuning (Sec. \ref{sect: ft}), we apply a clustering algorithm (such as K-Means) to identify clusters based on the content features learned from pre-training. That is, we do not perform supervised fine-tuning of the features. After finding the clusters, we calculate the centroid of each cluster (instead of each class) according to Eq. \ref{eq: centroid}. Centroids of the clusters (instead of the classes) are then used for OSR as discussed in Sec. \ref{sec: osr}. For reference convenience, we assign an ID to each cluster. Unsupervised OSR outputs one of the known cluster IDs or unknown.
\section{experiments}
\label{sec: exp}

We evaluate the proposed feature decoupling approach with two types of fine-tuning functions as mentioned in Section \ref{sect: ft}: classification loss (cross-entropy loss) and representation loss (triplet loss). Moreover, to show that our proposed approach works on different datasets. We test the proposed approach on images and malware datasets. A web link will be provided for our implementation and datasets in the paper if it is published.

\noindent\textbf{Fashion-MNIST} \cite{xiao2017/online} is associated with 10 classes of clothing images. It contains 60,000 training and 10,000 testing examples. In the Fashion-MNIST dataset, each example is a 28x28 grayscale image. To simulate an open-set dataset, we randomly pick six digits as the known classes, while the rest are treated as the unknown class for testing.

\noindent\textbf{CIFAR-10}  \cite{krizhevsky2009learning} contains 60,000 32x32 color images in 10 classes, with 6,000 images per class. There are 50,000 training images and 10,000 test images. As the Fashion-MNIST datasets and the FCGs datasets only have one channel, for consistency, we first convert the color images to grayscale and randomly pick six classes out of the ten classes as the known classes. In contrast, the remaining classes are treated as the known class only existing in the test set. 

\noindent \textbf{Microsoft Challenge (MS)} \cite{DBLP:journals/corr/abs-1802-10135} contains disassembled malware samples from 9 families:``Ramnit'', ``Lollipop'', ``Kelihos ver3'', ``Vundo'', ``Simda'', ``Tracur'', ``Kelihos ver1'', ``Obfuscator.ACY '' and ``Gatak''. We use 10260 samples that can be correctly parsed then extracted their FCGs as in \cite{DBLP:conf/codaspy/HassenC17} for the experiment. We randomly pick six classes of digits as the known classes participant in the training, while the rest are considered as unknowns that only exist in the test set.

\noindent \textbf{Android Genome (AG)} consists of 1,113 benign android apps and 1,200 malicious android apps. Our colleague provides the benign samples, and the malicious samples are from \cite{zhou_jiang}. We select nine families with a relatively larger size for the experiment to be fairly split into the training set and the test set. The nine families contain 986 samples in total. We first use \cite{DBLP:conf/ccs/GasconYAR13} to extract the function instructions and then generated the FCGs as in \cite{DBLP:conf/codaspy/HassenC17}. Also, to simulate an open-set scenario, we randomly pick six digits as the known classes while considering the rest as the unknown class.

\subsection{Implementation details and comparison methods}
 Our proposed training process consists of two stages. In the first stage, we compare our proposed self-supervised feature decoupling (FD) approach with other self-supervised learning approaches: RotNet \cite{DBLP:conf/iclr/GidarisSK18}, Barlow Twins \cite{DBLP:conf/icml/ZbontarJMLD21}, DTAE \cite{DBLP:conf/icann/JiaC22}\cite{DBLP:journals/corr/abs-2205-06918}. We construct a fine-tuning network in the second stage to refine the learned content features. We experiment with classification loss (cross-entropy loss: ce) and representation loss (triplet loss: triplet) as loss functions in the fine-tuning network. Furthermore, To demonstrate that our proposed approach is effective for OSR problems, we compare our approach with OpenMax\cite{bendale2016towards}.
 
\subsubsection{Self-supervised feature decoupling}
 As illustrated in Figure \ref{fig: pre-train}, the pre-training stage includes a transformation module to facilitate the pre-text task. For the image datasets, we use the rotation of a multiplier of 90 degrees (e.g., 0, 90, 180, 270 degrees) in the transformation module. As for the malware datasets. We first extract the FCGs of each sample, then apply FCG-random \cite{DBLP:journals/corr/abs-2205-06918} on the FCGs. The transformed samples are then passed through an encoder. The padded input layer size varies for different datasets. For the Fashion-MNIST dataset, the input images are of size (28, 28) and are padded to get the size (32, 32) with one channel. For the CIFAR-10 dataset, the padded input size is (36, 36). For the FCG datasets (MS and Android), the padded input layer is in the size of (67, 67). The padded input layer is then flowed by two non-linear convolutional layers with 32 and 64 nodes. We apply the max-polling layers with kernel size (3, 3) and strides (2, 2). We also add batch normalization after each convolutional layer to complete the convolutional block. After the convolutional block, we use two fully connected non-linear layers with 256 and 128 hidden units for the image datasets Fashion-MNIST and CIFAR-10. We only use one fully connected non-linear layer with 256 hidden units for the graph dataset. Furthermore, the size of representations is nine dimensions for all the datasets in our experiments, with six dimensions for the content features and the remaining three for the non-content transformation features. The six-dimensional content features are connected to a decoder, which is simply the reverse of the encoder. The three-dimensional transformation features are further connected to a linear layer and then fed to a softmax layer for the transformation classification. We use the Relu activation function and set the Dropout's keep probability as 0.2. We use
Adam as the optimizer with a learning rate of 0.001. 

The comparison methods RotNet, Barlow Twins, and DTAE share the same backbone encoder architecture as our proposed method. Their representation layers have six dimensions. Also, we have generalized the original RotNet and Barlow Twins methods for fair comparison in our experiments. Specifically, the pre-text task of original RotNet was classifying the rotation degrees in \cite{DBLP:conf/iclr/GidarisSK18}, which is only applicable for the image datasets. Here, to make RotNet feasible for the FCG datasets, we extend the pre-text task to predicting the FCG-random transformation labels for the FCG datasets. Moreover, the original Barlow Twins impose constraints on the cross-correlation matrix between the representations of two transformed views in \cite{DBLP:conf/icml/ZbontarJMLD21}. Here, we apply the same constraints to the cross-correlation matrices between the transformed views and their corresponding original samples.

\subsubsection{Supervised fine-tuning}
In the fine-tuning network, the encoder and representation layer maintains the same architectures as the pre-trained network. Then, instead of connecting the representation layer to a decoder and a transformation classifier, we only connect the content features of the representation layer to a decision layer (for classification loss) in Figure \ref{fig: fine-tune-cls} or a representation loss function as shown in Figure \ref{fig: fine-tune-rep}. Likewise, for the comparison methods, we connect their representation layers to the decision layer (for classification loss) or a representation loss function.

As one of the comparison methods, OpenMax does not have a pre-training stage. It shares the same encoder architecture as our backbone network, then the representation is directly fed to a softmax layer.

\subsection{Evaluation criteria}

\begin{table*}[t]
\caption{The average ROC AUC scores of 30 runs at 100\% and 10\% FPR of OpenMax and a group of 5 methods (without pre-training, pre-training with RotNet, Barlow Twins, DTAE and Feature Decoupling (FD)) for two loss functions: cross-entropy loss and triplet loss under supervised OSR scenario. The values in bold are the highest values in each group.}
\centering
\resizebox{0.93\textwidth}{!}{%
\begin{tabular}{l l cc cc cc cc cc cc}
\toprule
 & \multicolumn{1}{c}{}  & \multicolumn{2}{c}{Fashion-MNIST}  & \multicolumn{2}{c}{CIFAR-10} & \multicolumn{2}{c}{MS} & \multicolumn{2}{c}{AG} \\ 
  & FPR & 100\% & 10\%  & 100\% & 10\% & 100\% & 10\% & 100\% & 10\%\\ \cmidrule(l){3-4} \cmidrule(l){5-6} \cmidrule(l){7-8} \cmidrule(l){9-10}  
 \multirow{1}{*}{OpenMax} &  & 0.740\tiny{$\pm$0.046} & 0.016\tiny{$\pm$0.008} & 0.675\tiny{$\pm$0.017} & 0.006\tiny{$\pm$0.001} & 0.880\tiny{$\pm$0.037} & 0.040\tiny{$\pm$0.002} & 0.480\tiny{$\pm$0.190} & 0.001\tiny{$\pm$0.001}\\ \midrule
\multirow{5}{*}{ce}   & No Pre-training & 0.717\tiny{$\pm$0.036} & 0.029\tiny{$\pm$0.005} & 0.580\tiny{$\pm$0.046} & 0.007\tiny{$\pm$0.001} & 0.914\tiny{$\pm$0.030} & 0.052\tiny{$\pm$0.006} & 0.853\tiny{$\pm$0.082} & 0.022\tiny{$\pm$0.014}\\ 
& RotNet & 0.736\tiny{$\pm$0.047} & 0.031\tiny{$\pm$0.007} & 0.612\tiny{$\pm$0.040} & 0.008\tiny{$\pm$0.001} & 0.911\tiny{$\pm$0.032} & 0.055\tiny{$\pm$0.005} & 0.870\tiny{$\pm$0.059} & \textbf{0.026}\tiny{$\pm$0.017}\\
& Barlow Twins & 0.719\tiny{$\pm$0.034} & 0.028\tiny{$\pm$0.007} & 0.606\tiny{$\pm$0.017} & 0.007\tiny{$\pm$0.001} & 0.915\tiny{$\pm$0.022} & 0.053\tiny{$\pm$0.003} & 0.850\tiny{$\pm$0.068} & 0.020\tiny{$\pm$0.011}\\
& DTAE & 0.748\tiny{$\pm$0.040} & 0.032\tiny{$\pm$0.006} & 0.618\tiny{$\pm$0.019} & 0.008\tiny{$\pm$0.001} & 0.941\tiny{$\pm$0.018} & \textbf{0.064}\tiny{$\pm$0.002} & 0.855\tiny{$\pm$0.079} & 0.023\tiny{$\pm$0.013}\\
& FD (ours) & \textbf{0.771}\tiny{$\pm$0.032} & \textbf{0.034}\tiny{$\pm$0.006} & \textbf{0.628}\tiny{$\pm$0.012} & \textbf{0.009}\tiny{$\pm$0.001} & \textbf{0.945}\tiny{$\pm$0.010} & 0.060\tiny{$\pm$0.002} & \textbf{0.876}\tiny{$\pm$0.047} & 0.025\tiny{$\pm$0.013} \\
\midrule
\multirow{5}{*}{triplet} & No Pre-training & 0.716\tiny{$\pm$0.037} & 0.021\tiny{$\pm$0.005} & 0.610\tiny{$\pm$0.026} & 0.008\tiny{$\pm$0.001} & 0.923\tiny{$\pm$0.028} & 0.056\tiny{$\pm$0.005} & 0.868\tiny{$\pm$0.046} & 0.027\tiny{$\pm$0.014}\\ 
& RotNet & 0.743\tiny{$\pm$0.028} & \textbf{0.025}\tiny{$\pm$0.005} & 0.628\tiny{$\pm$0.015} & 0.009\tiny{$\pm$0.001} & 0.924\tiny{$\pm$0.018} & 0.057\tiny{$\pm$0.003} & 0.870\tiny{$\pm$0.036} & 0.025\tiny{$\pm$0.009}\\
& Barlow Twins & 0.709\tiny{$\pm$0.041} & 0.021\tiny{$\pm$0.007} & 0.621\tiny{$\pm$0.016} & 0.009\tiny{$\pm$0.001} & 0.918\tiny{$\pm$0.018} & 0.054\tiny{$\pm$0.003} & 0.871\tiny{$\pm$0.035} & 0.022\tiny{$\pm$0.006}\\ 
& DTAE & 0.744\tiny{$\pm$0.028}& 0.023\tiny{$\pm$0.003} & 0.632\tiny{$\pm$0.015} & 0.009\tiny{$\pm$0.001} & 0.928\tiny{$\pm$0.017} & \textbf{0.061}\tiny{$\pm$0.002} &\textbf{0.879}\tiny{$\pm$0.030} & \textbf{0.026}\tiny{$\pm$0.010}\\
& FD (ours) & \textbf{0.758}\tiny{$\pm$0.030} & \textbf{0.025}\tiny{$\pm$0.004} & \textbf{0.636}\tiny{$\pm$0.016} & \textbf{0.010}\tiny{$\pm$0.001} & \textbf{0.941}\tiny{$\pm$0.014} & \textbf{0.061}\tiny{$\pm$0.003} & 0.876\tiny{$\pm$0.029} & 0.025\tiny{$\pm$0.011}\\
\bottomrule
\end{tabular}}
\label{tab:auc}
\end{table*}

\begin{table*}
\caption{The average F1 scores of 30 runs OpenMax and a group of 5 methods (without pre-training, pre-training with RotNet, Barlow Twins, DTAE and Feature Decoupling) for two loss functions (cross entropy loss and triplet loss) under supervised OSR scenario. The values are the highest values in each group.}
\centering
\resizebox{0.8\textwidth}{!}{%
\begin{tabular}{l l ccc ccc ccc ccc}
\toprule
 Image Dataset& \multicolumn{1}{c}{}  & \multicolumn{3}{c}{Fashion-MNIST}  & \multicolumn{3}{c}{CIFAR-10} \\
  &  & Known & Unknown & Overall & Known & Unknown & Overall \\  \cmidrule(l){3-5} \cmidrule(l){6-8} 
 \multirow{1}{*}{OpenMax} &  & 0.747{\tiny$\pm$0.049} & 0.521{\tiny$\pm$0.178} & 0.714{\tiny$\pm$0.051} & 0.645{\tiny$\pm$0.022} & 0.540{\tiny$\pm$0.065} & 0.630{\tiny$\pm$0.017}\\ \midrule
\multirow{5}{*}{ce}   & No Pre-training & 0.685{\tiny$\pm$0.102} & 0.559{\tiny$\pm$0.076} & 0.667{\tiny$\pm$0.086} & 0.567\tiny{$\pm$0.048} & 0.369\tiny{$\pm$0.169} & 0.538\tiny{$\pm$0.045}\\ 
& RotNet & 0.711\tiny{$\pm$0.067} & 0.569\tiny{$\pm$0.097} & 0.691\tiny{$\pm$0.058} & 0.561\tiny{$\pm$0.061} & 0.472\tiny{$\pm$0.136} & 0.548\tiny{$\pm$0.049}\\
& Barlow Twins & 0.738{\tiny$\pm$0.025} & 0.506{\tiny$\pm$0.066} & 0.704{\tiny$\pm$0.025} & \textbf{0.599}\tiny{$\pm$0.022} & 0.395\tiny{$\pm$0.105} & 0.570\tiny{$\pm$0.023}\\
& DTAE & 0.733{\tiny$\pm$0.050} & 0.570{\tiny$\pm$0.087} & 0.710{\tiny$\pm$0.041} & 0.591\tiny{$\pm$0.037} & 0.472\tiny{$\pm$0.096} & 0.574\tiny{$\pm$0.027}\\
& FD (ours) & \textbf{0.748}\tiny{$\pm$0.024} & \textbf{0.587}\tiny{$\pm$0.075} & \textbf{0.725}\tiny{$\pm$0.022} & 0.587\tiny{$\pm$0.031} & \textbf{0.514}\tiny{$\pm$0.067} & \textbf{0.576}\tiny{$\pm$0.025} \\
\midrule
\multirow{5}{*}{triplet} & No Pre-training & 0.749\tiny{$\pm$0.014} & 0.505\tiny{$\pm$0.075} & 0.714\tiny{$\pm$0.021} & 0.579\tiny{$\pm$0.042} & 0.451\tiny{$\pm$0.134} & 0.561\tiny{$\pm$0.038}\\ 
& RotNet & 0.751\tiny{$\pm$0.015} & 0.537\tiny{$\pm$0.075} & 0.720\tiny{$\pm$0.020} & 0.603\tiny{$\pm$0.033} & 0.497\tiny{$\pm$0.087} & 0.588\tiny{$\pm$0.030} \\
& Barlow Twins &  0.740\tiny{$\pm$0.015} & 0.433\tiny{$\pm$0.042} & 0.696\tiny{$\pm$0.016} & 0.609\tiny{$\pm$0.025} & 0.446\tiny{$\pm$0.104} & 	0.586\tiny{$\pm$0.026}\\ 
& DTAE & \textbf{0.755}\tiny{$\pm$0.010} & 0.545\tiny{$\pm$0.076} & 0.725\tiny{$\pm$0.018} & \textbf{0.620}\tiny{$\pm$0.027} & 0.472\tiny{$\pm$0.086} & 0.599\tiny{$\pm$0.028}\\
& FD (ours) & 0.753\tiny{$\pm$0.011} & \textbf{0.582}\tiny{$\pm$0.083} & \textbf{0.729}\tiny{$\pm$0.018} & 0.617\tiny{$\pm$0.030} & \textbf{0.515}\tiny{$\pm$0.028} & \textbf{0.603}\tiny{$\pm$0.026}\\
\midrule \midrule
Malware Dataset& \multicolumn{1}{c}{}  &  \multicolumn{3}{c}{MS} & \multicolumn{3}{c}{AG} \\
  &  & Known & Unknown & Overall & Known & Unknown & Overall \\  \cmidrule(l){3-5} \cmidrule(l){6-8} 
 \multirow{1}{*}{OpenMax} &  & 0.891\tiny{$\pm$0.006} & 0.737\tiny{$\pm$0.010} & 0.869\tiny{$\pm$0.006} & 0.408\tiny{$\pm$0.190} & 0.640\tiny{$\pm$0.163} & 0.441\tiny{$\pm$0.184}\\ \midrule
\multirow{5}{*}{ce}   & No Pre-training & 0.899\tiny{$\pm$0.010} & 0.703\tiny{$\pm$0.061} & 0.871\tiny{$\pm$0.017} & 0.683\tiny{$\pm$0.117} & 0.540\tiny{$\pm$0.329} & 0.663\tiny{$\pm$0.120}\\ 
& RotNet & 0.900\tiny{$\pm$0.012} & 0.708\tiny{$\pm$0.077} & 0.872\tiny{$\pm$0.021} & 0.709\tiny{$\pm$0.121} & \textbf{0.613}\tiny{$\pm$0.335} & 0.695\tiny{$\pm$0.135}\\
& Barlow Twins & 0.896\tiny{$\pm$0.007} & 0.712\tiny{$\pm$0.039} & 0.870\tiny{$\pm$0.011} & 0.701\tiny{$\pm$0.093} & 0.541\tiny{$\pm$0.309} & 0.678\tiny{$\pm$0.113}\\
& DTAE & \textbf{0.908}\tiny{$\pm$0.008} & \textbf{0.779}\tiny{$\pm$0.027} & \textbf{0.890}\tiny{$\pm$0.010} & 0.686\tiny{$\pm$0.107} & 0.535\tiny{$\pm$0.280} & 0.664\tiny{$\pm$0.110} \\
& FD (ours) & 0.905\tiny{$\pm$0.007} & 0.771\tiny{$\pm$0.026} & 0.886\tiny{$\pm$0.009} & \textbf{0.711}\tiny{$\pm$0.096} & 0.612\tiny{$\pm$0.339} & \textbf{0.697}\tiny{$\pm$0.118}\\
\midrule
\multirow{5}{*}{triplet} & No Pre-training & 0.905\tiny{$\pm$0.007} & 0.728\tiny{$\pm$0.035} & 0.879\tiny{$\pm$0.011} & 0.753\tiny{$\pm$0.074} & 0.789\tiny{$\pm$0.133} & 0.758\tiny{$\pm$0.068} \\ 
& RotNet & 0.906\tiny{$\pm$0.008} & 0.739\tiny{$\pm$0.031} & 0.882\tiny{$\pm$0.011} & 0.755\tiny{$\pm$0.069} & 0.791\tiny{$\pm$0.178} & 0.760\tiny{$\pm$0.074}\\
& Barlow Twins & 0.896\tiny{$\pm$0.006} & 0.699\tiny{$\pm$0.034} & 0.868\tiny{$\pm$0.010} & \textbf{0.761}\tiny{$\pm$0.081} & 0.739\tiny{$\pm$0.247} & 0.757\tiny{$\pm$0.091}\\ 
& DTAE & \textbf{0.911}\tiny{$\pm$0.006} & 0.751\tiny{$\pm$0.024} & \textbf{0.889}\tiny{$\pm$0.009} & 0.734\tiny{$\pm$0.079} & 0.735\tiny{$\pm$0.197} & 0.734\tiny{$\pm$0.081}\\
& FD (ours) & 0.909\tiny{$\pm$0.007} & \textbf{0.762}\tiny{$\pm$0.031} & \textbf{0.889}\tiny{$\pm$0.010} & 0.760\tiny{$\pm$0.059} & \textbf{0.807}\tiny{$\pm$0.160} & \textbf{0.766}\tiny{$\pm$0.061}\\
\bottomrule
\end{tabular}}
\label{tab:f1}
\end{table*}

To simulate an open-set scenario, we randomly pick six classes as the known classes and use them in the training process. The ensemble of the remaining classes is considered the unknown class, which does not participate in the training process and only exists in the test set. We experiment with both supervised and unsupervised OSR scenarios. 

For the supervised scenario, we simulate three groups of such open sets and experiment with each group with ten runs. We calculate the average results of these 30 runs when evaluating the model performances. For evaluation, we perform a three-dimensional comparison of our proposed approach. First, we compare model performances with and without using the pre-training process to verify that the pre-training process benefits the OSR problem for different loss functions. Second, we compare our feature decoupling (FD) approach with other self-supervised pre-training approaches, RotNet, Barlow Twins, and DTAE. Finally, to show that the two-stage trained model can achieve good performance compared to other OSR approach, we compare the proposed approach with the popular OSR solution OpenMax. Similar to the unsupervised scenario, we measure both ROC AUC scores under 100\% and 10\% FPRs. The ROC AUC scores under 100\% FPR is commonly used in measuring model performance. However, in real-life applications such as malware detection, a lower FPR is more desirable. Thus the ROC AUC scores under 10\% FPR are more meaningful in these cases. Moreover, as the objective of the OSR problem is twofold: classifying the known classes and recognizing the unknown class, we evaluate the F1 scores for the known class and the unknown class separately.

\begin{figure*}
\centering
\begin{subfigure}[t]{.215\textwidth}
    \includegraphics[trim={0 0 2.15cm 0},clip, width=\linewidth, height=0.16\textheight]{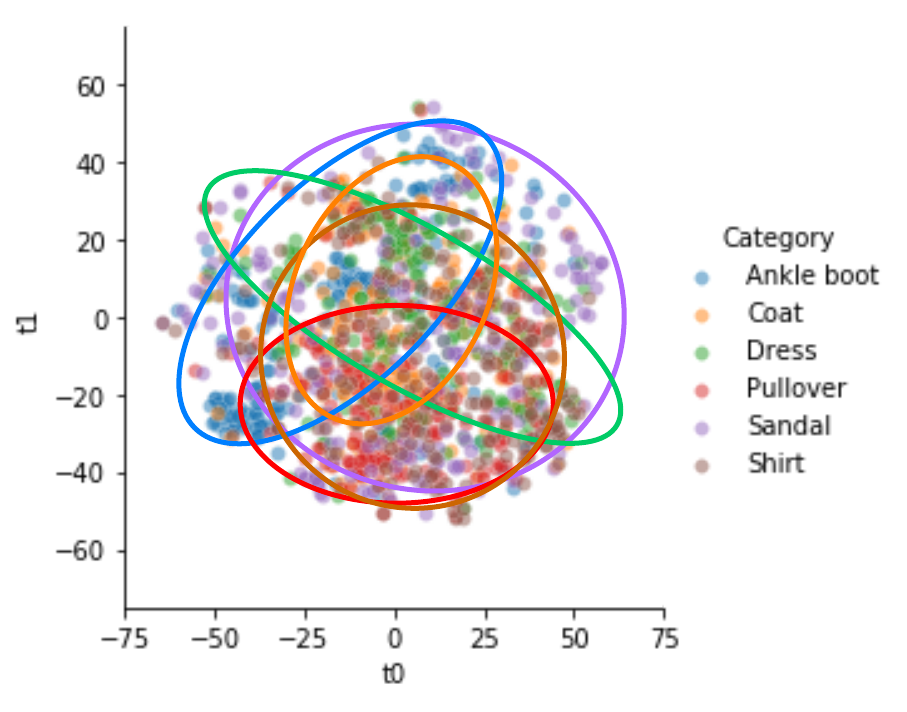}
    \caption{RotNet}
\label{fig: hm1}
\end{subfigure}
\begin{subfigure}[t]{.215\textwidth}
    \includegraphics[trim={0 0 2.15cm 0},clip,width=\linewidth, height=0.16\textheight]{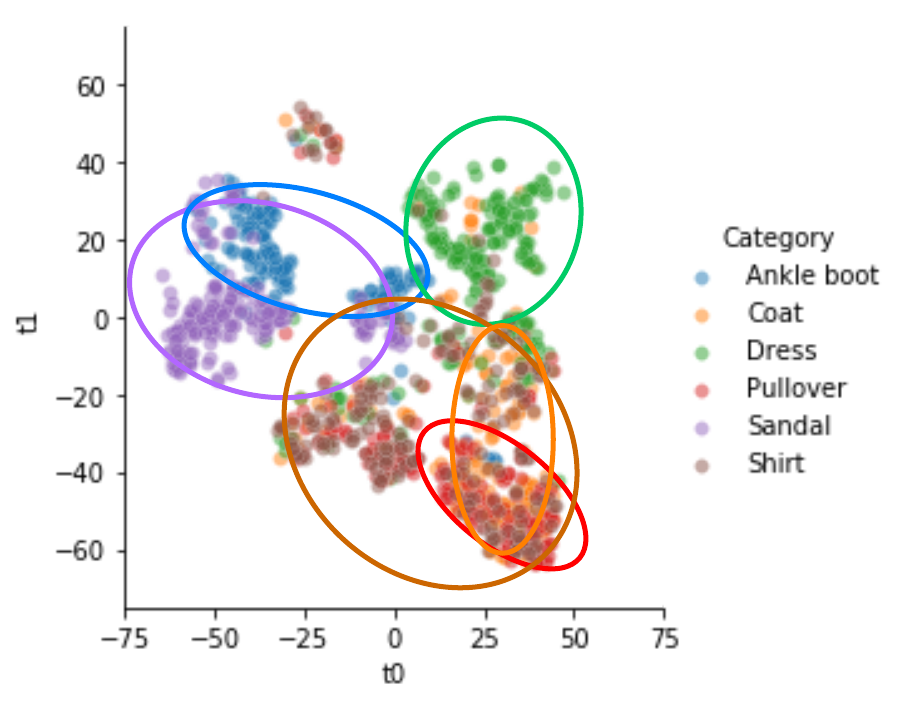}
    \caption{Barlow Twins}
\label{fig: hm2}
\end{subfigure}
\begin{subfigure}[t]{.215\textwidth}
    \includegraphics[trim={0 0 2.15cm 0},clip,width=\linewidth, height=0.16\textheight]{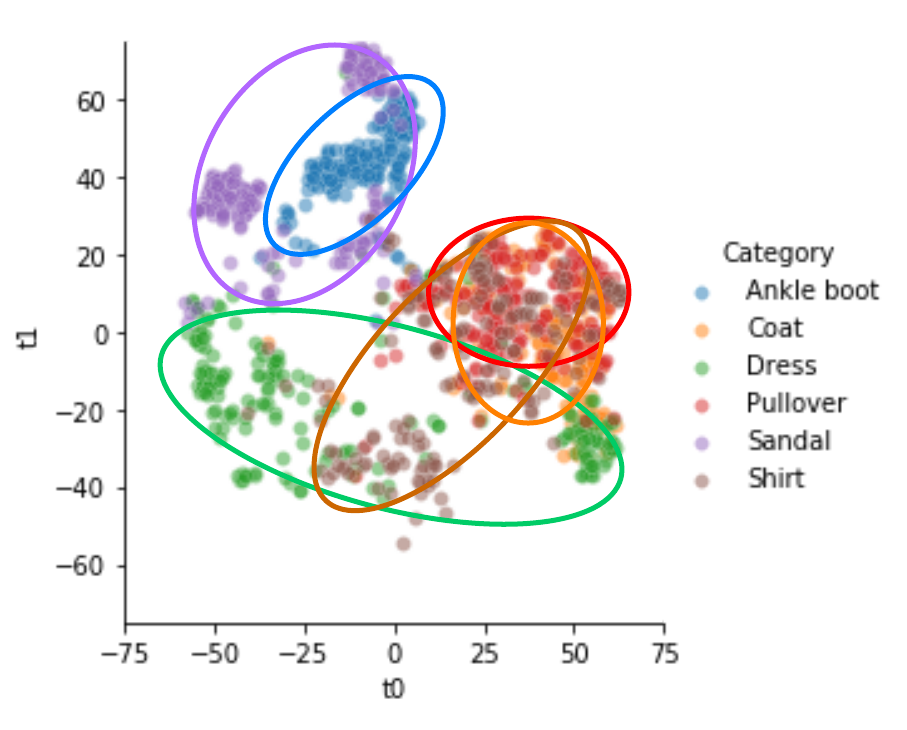}
    \caption{DTAE}
\label{fig: hm3}
\end{subfigure}
\begin{subfigure}[t]{.3\textwidth}
    \includegraphics[width=\linewidth, height=0.16\textheight]{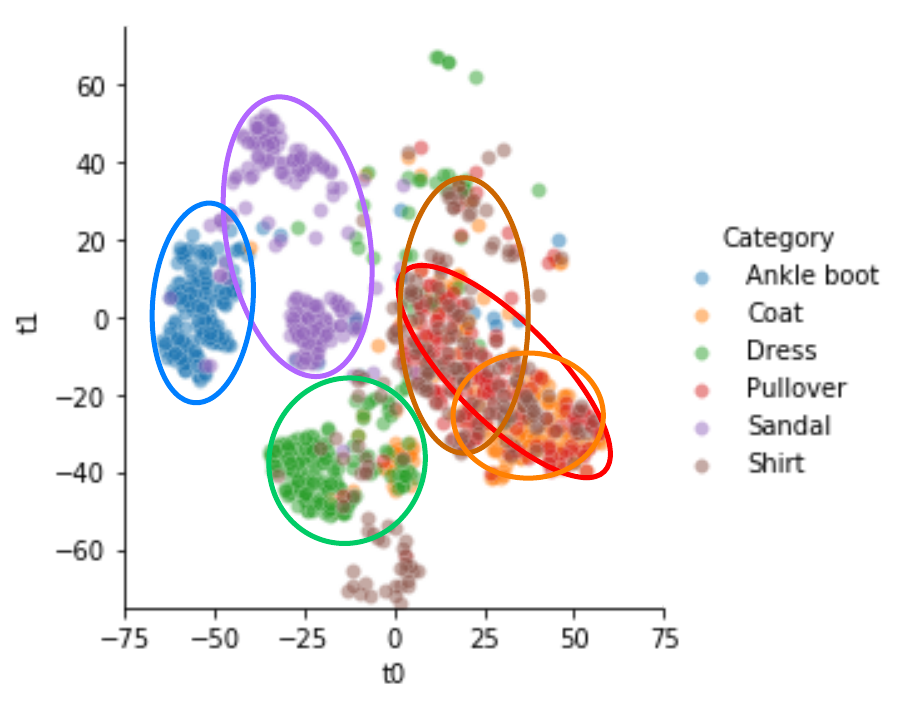}
    \caption{Feature Decoupling}
\label{fig: hm4} 
\end{subfigure}

\caption{The t-SNE plots of the representations of Fashion-MNIST test samples on pre-trained models.}
\label{fig: pre-train-tsne}
\end{figure*}



\begin{figure}
    \includegraphics[width=\linewidth]{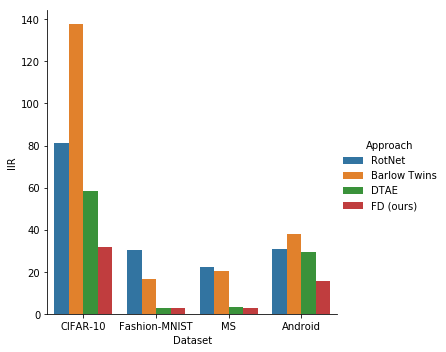}
\caption{IIR after the pre-training stage.}
\label{fig: iir}
\end{figure}

\subsection{Evaluation results}

Table \ref{tab:auc} reports the AUC ROC scores under different FPR values: 100\% and 10\% under the supervised OSR scenario, where the known class labels are available. Comparing the ``No Pre-training'' rows with ``OpenMax" rows of both loss functions, we observe that without the pre-training stage, OpenMax outperforms the cross-entropy loss and triplet loss in the image datasets. On the contrary, for the malware datasets, the cross-entropy and triplet loss perform better than OpenMax. Moreover, comparing the models without pre-training stages with those with pre-training stages, we observe that the pre-training methods benefit the model performances in most cases. Also, the model pre-trained with our proposed feature decoupling (FD) achieves the best performance in 12 out of 16 comparison groups (4 datasets x 2 FPRs x 2 loss functions). Especially, the model pre-trained with our proposed approach achieves the best performance in all the cases in the graph datasets.

Besides the AUC ROC scores, we measure the F1 scores of different methods in Table \ref{tab:f1}. Notably, we measure the performance under three categories: the average F1 scores of all the known class ("Known" columns), the F1 scores of the unknown class ("Unknown" columns), and the average F1 scores of the known and unknown classes ("Overall" columns). Similar to the AUC ROC results, OpenMax outperforms cross-entropy loss and triplet loss in the image datasets when no pre-training stage is involved. However, both loss functions surpass OpenMax in the malware datasets. Moreover, all the pre-training methods benefit the model performance in classifying the known classes and recognizing the unknown class in most cases. Our proposed approach outperforms the other pre-training methods in 15 out of 24 groups (4 datasets x 3 categories x 2 loss functions). Especially for the "Overall" performances, our proposed approach achieves the best performance in 7 out of 8 groups.

From the experiment results, we observe that the performance of OpenMax differs on image and malware datasets. Also, a pre-training stage boosts the model performance on both classification and representation loss. Moreover, in most cases, our proposed self-supervised feature decoupling approach outperforms the other pre-training methods.

We perform an ablation study from two perspectives for our approach. First, from the two-stage training perspective, we study the effects of the pre-training stage. We compare the AUC scores in the ``No Pre-training'' rows and ``FD(ours)'' rows in Table \ref{tab:auc} and Table \ref{tab:f1}. As expected, we observe that the pre-training stage has played an important role in the process. Second, our proposed pre-training approach, Feature Decoupling, has two components: a content reconstructor and a transformation classifier. The content reconstructor shares the same objective as DTAE. The results in the ``DTAE'' rows and ``FD(ours)'' rows in Table \ref{tab:auc} and Table \ref{tab:f1} indicate that the transformation classifier usually contributes to improved performance in our proposed approach.

\begin{figure*}[t]
 \begin{subfigure}[b]{0.49\textwidth}
               \includegraphics[width=\linewidth]{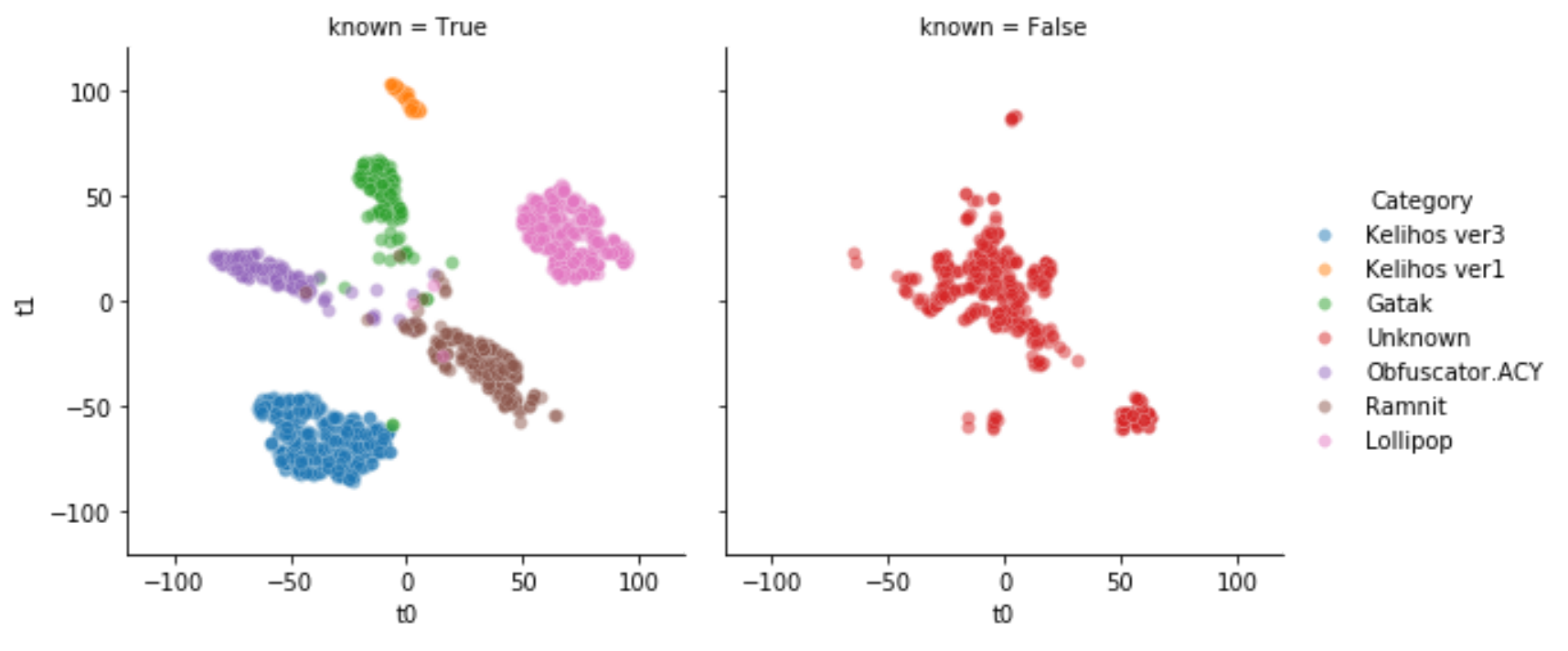}
                \caption{OpenMax}
                \label{fig: tsne-openmax}
        \end{subfigure}
 \begin{subfigure}[b]{0.49\textwidth}               \includegraphics[width=\linewidth]{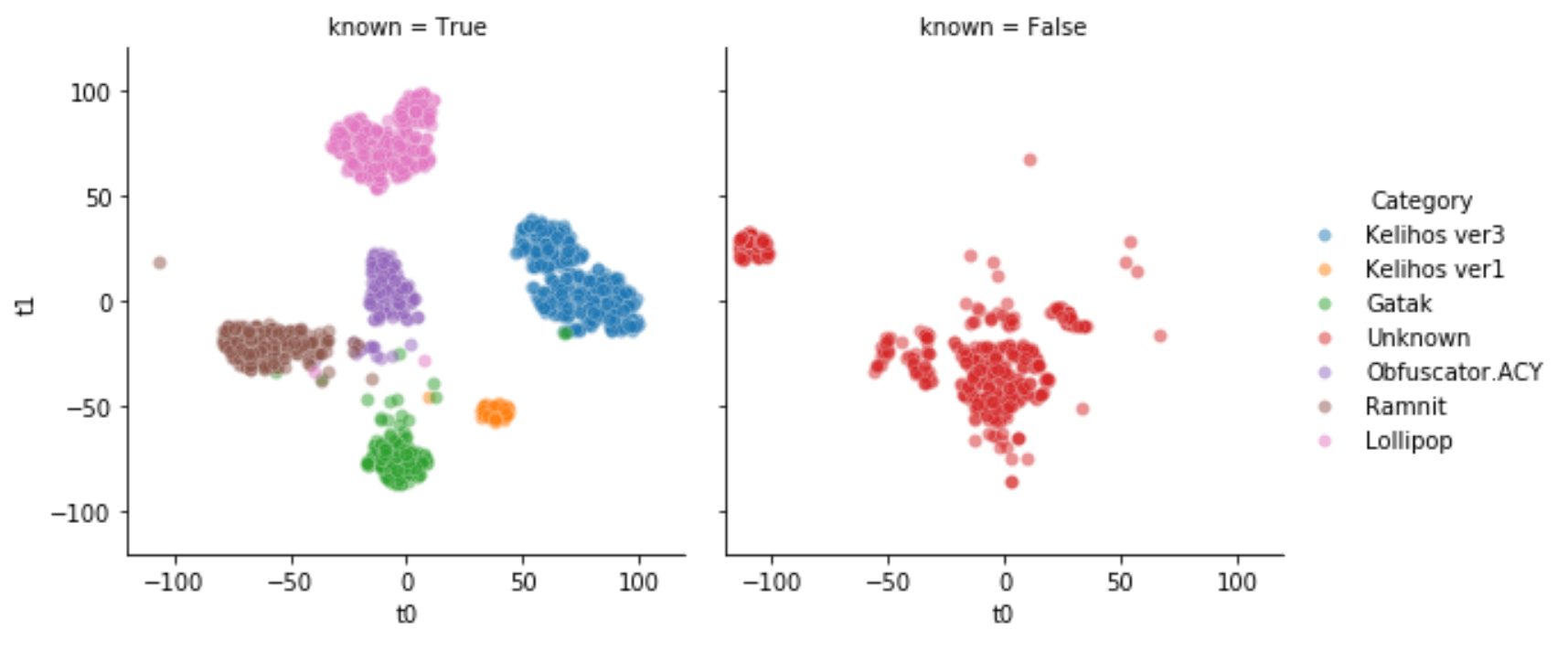}
                \caption{No Pre-training}
                \label{fig: tsne-no-pre}
        \end{subfigure}  \\
 \begin{subfigure}[b]{0.49\textwidth}
               \includegraphics[width=\linewidth]{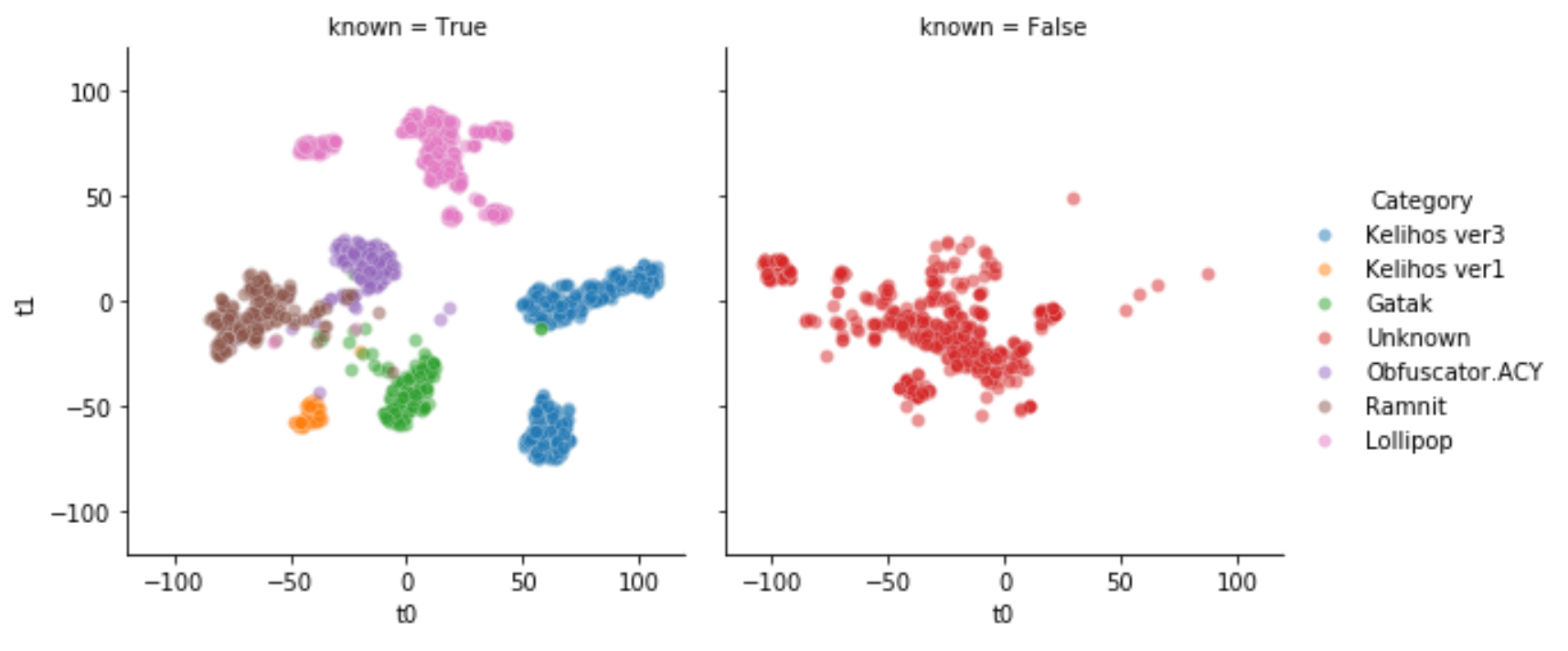}
                \caption{RotNet}
                \label{fig: tsne-rotnet}
        \end{subfigure} 
  \begin{subfigure}[b]{0.49\textwidth}                \includegraphics[width=\linewidth]{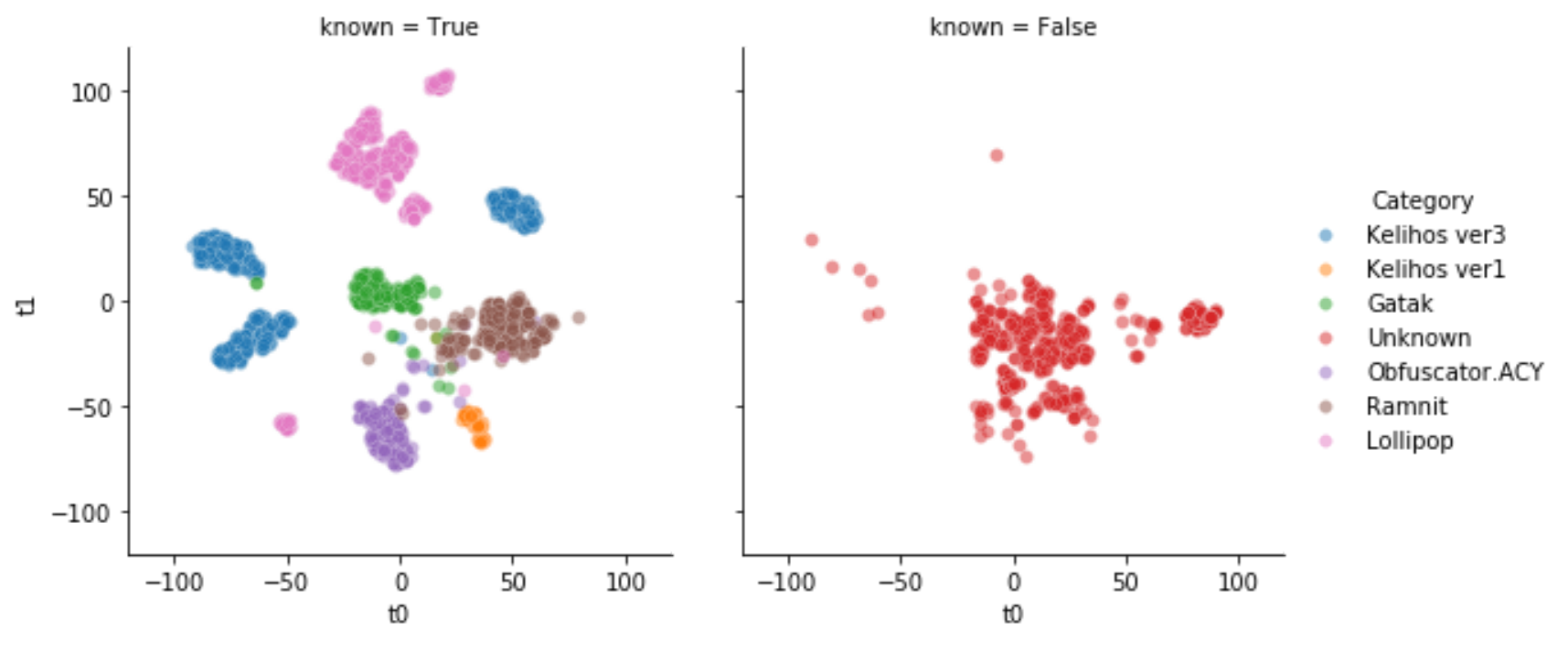}
                \caption{Barlow Twins}
                \label{fig: tsne-bt}
        \end{subfigure} \\
 \begin{subfigure}[b]{0.49\textwidth}
               \includegraphics[width=\linewidth]{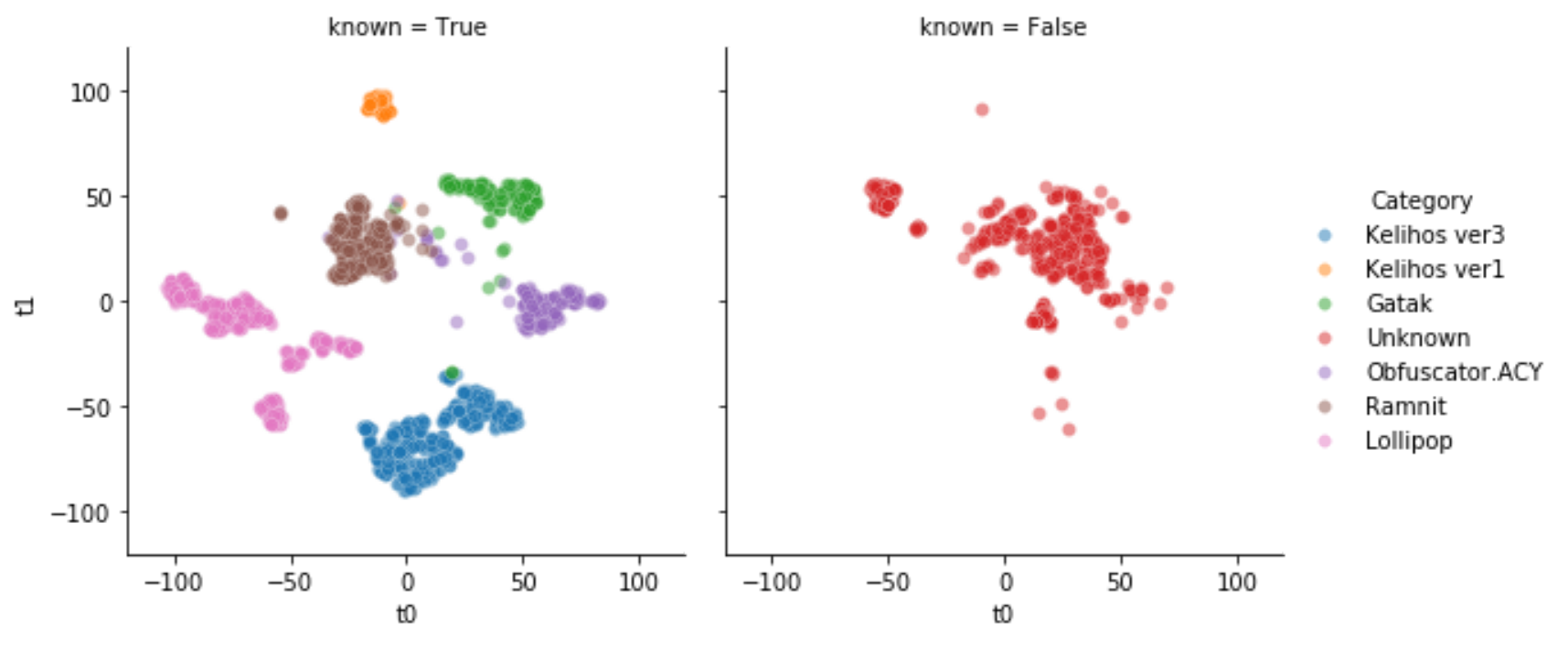}
                \caption{DTAE}
                \label{fig: tsne-dtae}
        \end{subfigure} 
  \begin{subfigure}[b]{0.49\textwidth}                \includegraphics[width=\linewidth]{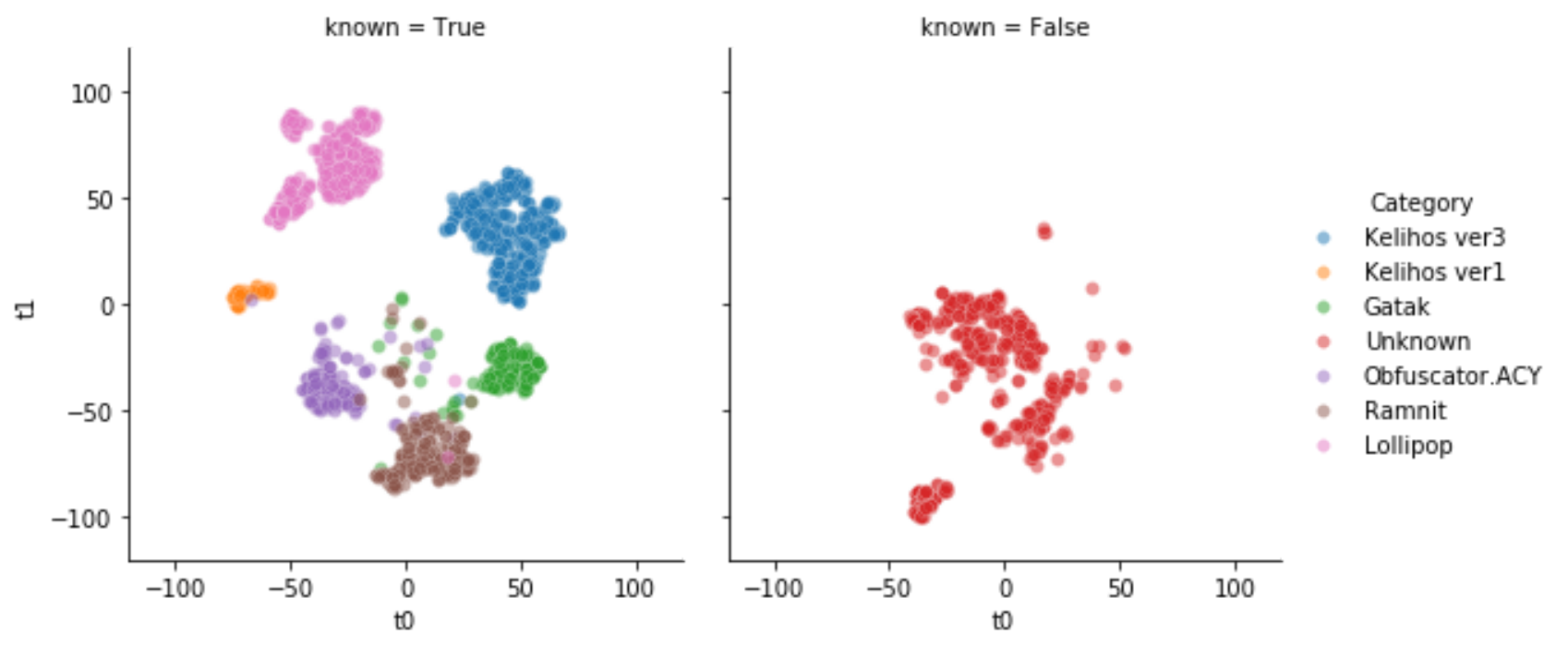}
                \caption{Feature Decoupling}
                \label{fig: tsne-fd}
        \end{subfigure}
\caption{The t-SNE plots of the representations of MS test samples learned by different models: (a) OpenMax; (b) triplet loss without pre-training; (c) triplet loss pre-trained with RotNet; (d) triplet loss pre-trained with Barlow Twins; (e) triplet loss pre-trained with DTAE; (f) triplet loss pre-trained with Feature Decoupling. The left subplots are the representations of the known class, and the right subplots are the representations of the unknown classes.}
\label{fig: fine-tuned-tsne}
\end{figure*}

\subsection{Analysis of the self-supervised models}
\label{sec: analysis-pre}
Our experiment results indicate that the proposed feature decoupling approach benefits different loss functions on the OSR tasks. We plot the t-SNE plots at different stages to further analyze the model performances. Figure \ref{fig: pre-train-tsne} shows the t-SNE plots of the known classes in the (unseen) test set of Fashion-MNIST after pre-training. 
The models are pre-trained by self-supervised learning approaches: RotNet, Barlow Twins, DTAE, and Feature Decoupling. Comparing the four approaches, we observe that RotNet fails to separate any of the six known classes, while the other three approaches manage to separate the known classes to some level. Among the other three approaches, Barlow Twins and Feature Decoupling can better cluster "Dress" samples, whereas the representations of the "Dress" samples learned by DTAE are more spread out and meanwhile overlap with the representations of the "Shirt" samples. Moreover, the representations of "Ankle boot" and "Sandal" samples learned by Feature Decoupling are more separable than the other approaches.

Besides visually evaluating representations via t-SNE plots, we propose intra-inter ratio (IIR) to measure the representation quality learned by different self-supervised pre-training approaches. For class $k$, we define the intra class spread as the average distance of instances from its centroid:

\begin{equation}
    intra_k = \frac{1}{N_k} \sum_{i=1}^{N_k} d( \mu_k, z_i),
\label{eq: intra}
\end{equation}
where $N_k$ is the number of samples in class $k$ and $d(.,.)$ is a distance function. Meanwhile, we measure the inter separation of the class $k$ as the distance of the its centroid $\mu_k$ to its nearest centroid of other classes:

\begin{equation}
    inter_k = \min_{i, i\neq k} d(\mu_k, \mu_i) 
\label{eq: inter}
\end{equation}

Moreover, the intra-inter ratio of class $k$ can be then defined as $IIR_k = intra_k / inter_k$, which combines both intra-spread and inter-separation for the representation quality measurement. Here, we use the average IIR over all the $C$ known classes to further measure the representation quality:
\begin{equation}
    IIR = \frac{1}{K} \sum_{k=1}^{K} \frac{intra_k}{inter_k}
\label{eq: iir}
\end{equation}

IIR is similar to the feature space density proposed by Roth et al.  \cite{DBLP:conf/icml/RothMSGOC20}. One difference is that IIR calculates the average ratio of all classes instead of the ratio of the average intra-distance and inter-distance. That is, IIR focuses on the representation quality of each class before considering the overall quality. Also, the inter-distance in IIR is calculated with respect to the nearest centroid, while in feature space density, it is an average of all pairs of centroids. That is, inter-distance in IIR is designed to characterize the "near miss" centroid that is most likely to cause misclassification. 

A lower IIR score indicates lower intra-spread and/or higher inter-separation, which characterizes better representation quality. Figure \ref{fig: iir} shows the IIR of different datasets after the pre-training stage. We observe that our proposed Feature Decoupling (FD) outperforms the other pre-training approaches for the image and malware datasets. Note that self-supervised pre-training does not use class labels, but FD can yield better representations in the t-SNE plots and lower IIR in the test set.





\subsection{Analysis of the fine-tuned models}

Besides the self-supervised learning stage, we also visualize the difference between learned representations after the fine-tuning stage under the supervised OSR scenario. Figure \ref{fig: fine-tuned-tsne} shows the representations of known and unknown MS malware samples learned by different methods after the fine-tuning stage. In these experiments, we consider ``Kelihos ver3'', ``Kelihos ver1'', ``Gatak'', ``Obfuscator.ACY'', ``Ramnit'' and ``Lollipop'' as the known classes, and the samples of the remaining three classes ``Vundo'', ``Simda'', ``Tracur'' together are treated as the unknown class not a participant in the training process. Figure \ref{fig: tsne-openmax} and Figure \ref{fig: tsne-no-pre} do not involve the pre-training stage. The model used in Figure \ref{fig: tsne-openmax} is trained by OpenMax, and Figure \ref{fig: tsne-no-pre} is trained by triplet loss directly. For comparison, the models in Figure \ref{fig: tsne-rotnet} - Figure \ref{fig: tsne-fd} are pre-trained by different self-supervision approaches and fine-tuned by triplet loss. From the representations of the unknown classes in the left subplots, we observe that OpenMax, triplet loss without pre-training, and Feature Decoupling perform better in the intra-class spread. At the same time, the models pre-trained by RotNet, Barlow Twins, and DTAE tend to spread one class into several clusters, such as ``Kelihos ver3'' and ``Lollipop''. Furthermore, comparing the representations of the unknown samples in the right subplots, the representations of the unknown class learned by models in the pre-training stage are more concentrated near the origin and tend to achieve better intra-class spread. Comparing the left and right subplots, we observe that compared with other approaches, the representations of the known classes have less overlap with those of the unknown class in the Feature Decoupling approach.

\begin{figure}
    \includegraphics[width=\linewidth]{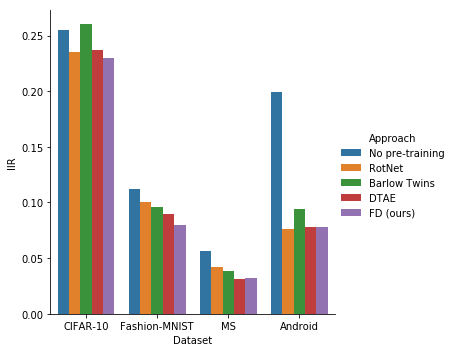}
\caption{IIR after the fine-tuning stage.}
\label{fig: iir-finetune}
\end{figure}

\begin{figure}
    \includegraphics[width=\linewidth]{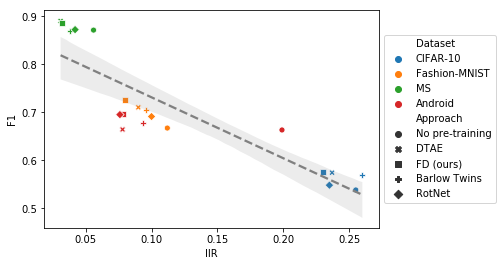}
\caption{F1 against IIR after the fine-tuning stage.}
\label{fig: f1-iir}
\end{figure}

Besides the triplet loss, we plot the IIR of the models fine-tuned by cross-entropy loss in Figure \ref{fig: iir-finetune}. Self-supervised pre-training benefits IIR in most cases, except for Barlow Twins in the CIFAR-10 dataset. Consistent with the IIR after the pre-training stage, the model pre-trained with Feature Decoupling benefits IIR in most cases. Furthermore, to determine if IIR can help explain OSR performance, we plot the overall F1 scores in Table \ref{tab:f1} against IIR in Figure \ref{fig: f1-iir}. We observe that F1 scores and IIR are highly correlated, where the Pearson correlation coefficient is -0.88. The strong correlation indicates that improvement in IIR can help explain enhancement in overall F1. Hence, self-supervised methods (such as FD) that can improve IIR can increase OSR performance.


\begin{figure}

\centering
\begin{subfigure}[t]{.49\columnwidth}
    \includegraphics[ width=\linewidth]{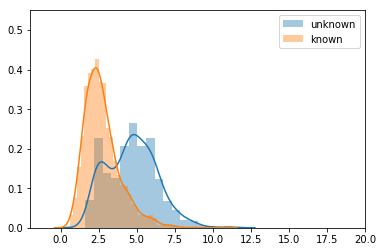}
    \caption{Without FD (ce)}
\label{fig: hist1}
\end{subfigure}
\begin{subfigure}[t]{.49\columnwidth}
    \includegraphics[ width=\linewidth]{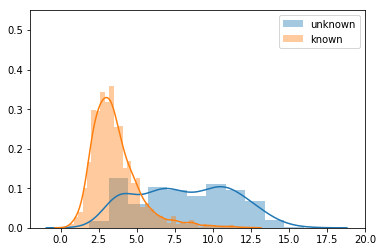}
    \caption{With FD (ce)}
\label{fig: hist2}
\end{subfigure}
\begin{subfigure}[t]{.49\columnwidth}
    \includegraphics[ width=\linewidth]{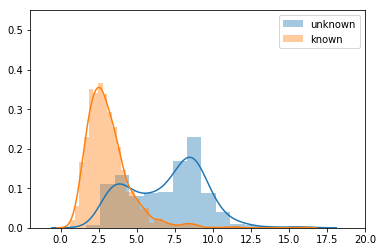}
    \caption{Without FD (triplet)}
\label{fig: hist3}
\end{subfigure}
\begin{subfigure}[t]{.49\columnwidth}
    \includegraphics[ width=\linewidth]{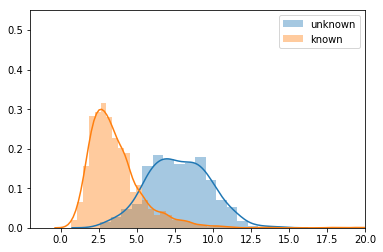}
    \caption{With FD (triplet)}
\label{fig: hist4}
\end{subfigure}
\caption{The distributions of outlier scores for the known and unknown classes of the MS dataset with and without our proposed Feature Decoupling (FD) pre-training process.}
\label{fig: hist}
\end{figure}

Figure \ref{fig: hist} shows the distributions of the outlier scores for the known and unknown classes in the MC dataset. Figure \ref{fig: hist1} and Figure \ref{fig: hist2} show the outlier scores distributions of the models using cross-entropy with and without Feature Decoupling pre-training, respectively. Figure \ref{fig: hist3} and Figure \ref{fig: hist4} show the outlier scores distributions of the models using triplet loss with and without Feature Decoupling pre-training, respectively. We observe that for both loss functions, the pre-training stage significantly increases outlier scores for the unknown class. Meanwhile, the outlier scores of the known classes are slightly increased. This effect pushes the outlier scores of the unknown class further away from the known classes and results in less overlap. The less overlap leads to higher accuracy of recognizing the unknown class.

\subsection{Experiments on unsupervised OSR}

We evaluate the unsupervised scenario discussed in Section \ref{sect: cluster} on the Fashion-MNIST and MS datasets. Although both datasets contain class labels, we only use the labels to create the open-set datasets and calculate model performance metrics. We perform K-Means (K=6) on the representations learned by the self-supervised models to find the potential class centroids. Similar to the supervised scenario, we simulate an open-set scenario by randomly picking six classes as the known classes. Also, we simulate three groups of such open sets and experiment with each group with three runs, resulting 9 runs. Then, we calculate the average results of these 9 runs when evaluating the model performances. We compare our feature decoupling (FD) approach with other self-supervised pre-training approaches, RotNet, Barlow Twins, and DTAE, and report the ROC AUC scores under 100\% and 10\% False Positive Rate (FPR) in Table \ref{tab:unsupervised-auc}. Our feature decoupling approach outperforms the other self-supervised learning approaches in both image and malware datasets. Though we expect AUC in the unsupervised OSR scenario (Table \ref{tab:unsupervised-auc}) to be lower than AUC in the supervised OSR scenario (Table \ref{tab:auc}), for feature decoupling (FD), the difference might not be huge. For Fashion-MINST, AUC with 100\% FPR in unsupervised OSR is 0.655, compared to 0.771 (ce) or 0.758 (triplet) in supervised OSR. This provides additional evidence (beyond Sec. \ref{sec: analysis-pre}) that the features learned via feature decoupling could be effective for OSR.

\begin{table}
\caption{The average ROC AUC scores of 9 runs at 100\% and 10\% FPR of a group of 4 methods (RotNet, Barlow Twins, DTAE and Feature Decoupling (FD)) for the unsupervised OSR scenario. The values in bold are the highest values.}
\centering
\resizebox{\columnwidth}{!}{%
\begin{tabular}{l cc cc cc cc}
\toprule
  \multicolumn{1}{c}{}  & \multicolumn{2}{c}{Fashion-MNIST}  & \multicolumn{2}{c}{MS} \\ 
   FPR & 100\% & 10\%  & 100\% & 10\% \\ \cmidrule(l){2-3} \cmidrule(l){4-5}
RotNet & 0.519\tiny{$\pm$0.089} & 0.005\tiny{$\pm$0.002} & 0.587\tiny{$\pm$0.063} & 0.008\tiny{$\pm$0.002} \\
Barlow Twins & 0.463\tiny{$\pm$0.059} & 0.004\tiny{$\pm$0.002} & 0.537\tiny{$\pm$0.120} & 0.007\tiny{$\pm$0.002} \\
DTAE & 0.639\tiny{$\pm$0.083} & 0.032\tiny{$\pm$0.006} & 0.639\tiny{$\pm$0.083} & 0.010\tiny{$\pm$0.001} \\
FD (ours) & \textbf{0.655}\tiny{$\pm$0.053} & \textbf{0.034}\tiny{$\pm$0.006} & \textbf{0.686}\tiny{$\pm$0.092} & \textbf{0.012}\tiny{$\pm$0.005} \\
\bottomrule
\end{tabular}}
\label{tab:unsupervised-auc}
\end{table}

\section{conclusion}
We use a two-stage learning approach for the OSR problems. We propose a self-supervised feature decoupling method to split the learned representation into the content and transformation parts in the first stage. In the second stage, we fine-tune the content features from the first stage with class labels. Furthermore, we consider an unsupervised OSR scenario, where we cluster the content features to find the potential classes in the second stage. We introduce intra-inter ratio (IIR) to evaluate the learned content representations. The results indicate that our feature decoupling method outperforms the other self-supervised learning methods in supervised and unsupervised OSR scenarios with image and malware datasets. Our analyses indicate that IIR is correlated with and can explain OSR performance.

\bibliographystyle{IEEEtran}
\bibliography{ref}

\end{document}